\pdfoutput=1
\documentclass[11pt]{article}

\usepackage{acl2023}
\usepackage{times}
\usepackage{latexsym}

\usepackage[T1]{fontenc}
\usepackage[utf8]{inputenc}
\usepackage{natbib}  % DO NOT CHANGE THIS AND DO NOT ADD ANY OPTIONS TO IT
\usepackage{caption} 
\usepackage{graphicx}
\usepackage{multirow,array}
\usepackage{bm}
\usepackage{bbm}
\usepackage{algorithm,algorithmicx}
\usepackage{amsmath}
\usepackage{amsthm}
\usepackage{verbatim}
\usepackage{booktabs}
\usepackage{amsfonts,amssymb} 
\usepackage[noend]{algpseudocode}

\newcommand{\secref}[1]{Section \ref{#1}}
\newcommand{\figref}[1]{Figure \ref{#1}}
\newcommand{\eqnref}[1]{Eq. (\ref{#1})}
\newcommand{\tabref}[1]{Table \ref{#1}}

\title{Pruning  Pre-trained Language Models with Principled Importance and Self-regularization}

\author{Siyu Ren \hspace*{1cm} Kenny Q. Zhu\textsuperscript{\rm}\thanks{\hspace{2mm}The corresponding author.}\\
	Shanghai Jiao Tong University\\
	Shanghai, China\\
	roy0702@sjtu.edu.cn, kzhu@cs.sjtu.edu.cn}
\begin{document}
	\maketitle
	\begin{abstract}
		 
Iterative pruning is one of the most effective compression methods for pre-trained language models. We discovered that finding the optimal pruning decision is an equality-constrained 0-1 Integer Linear Programming problem. 
The solution to this optimization problem leads to a principled importance criterion which we use to rank parameters during iterative model pruning.
To mitigate the poor generalization at high sparsity levels, we propose a self-regularization scheme 
where model prediction is regularized by the latest checkpoint with 
increasing sparsity throughout pruning. 
Our experiments on natural language understanding, 
question answering, named entity recognition, and data-to-text generation 
with various Transformer-based PLMs show the
effectiveness of the approach at various sparsity levels.
	\end{abstract}
	
	% Entries for the entire Anthology, followed by custom entries
	% \bibliography{anthology,custom}
	
%	
	\section{Introduction}

Pre-trained language models~(PLMs)~\cite{bert,gpt2} have significantly advanced the state-of-the-art in various natural language processing tasks~\cite{glue,webnlg,e2e,dart}. However, these models often contain a vast amount of parameters, posing non-trivial requirements for storage and computation. Due to this inefficiency, the applications of PLMs in resource-constrained scenarios are still limited.

To resolve the above challenge, model compression~\cite{pkd,svd,albert} has been actively studied to make PLMs meet the practical requirement. Among them, iterative pruning methods are widely adopted at only a tiny expense of model performance when adapting PLMs to downstream tasks. 
%These methods remove redundant parameters while updating remaining parameters using specific importance criteria to reduce model size effectively. 
During the course of iterative pruning, model parameters can not only be updated but also be pruned based on the rank of their importance scores in order to satisfy the cardinality constraint.
Prevalent importance criteria are based on the parameter's magnitude~\cite{gupta,mag2} or sensitivity~\cite{l0,movement,superticket,platon}. Parameters with low importance scores are pruned and are expected to have little impact on model performance.

Despite the empirical success, existing importance criteria for model pruning still face two major limitations: (1) they are heuristically defined and may not accurately quantify a parameter's contribution to the learning process, e.g., absolute weight value in magnitude-based pruning and gradient-weight product in sensitivity-based pruning; (2) they determine the importance of each parameter individually without considering the effect of coinstantaneous parameter updates on model performance, e.g., sensitivity is estimated by the absolute change in training error if only a single parameter is pruned and others remain unchanged.

%Though effective, such importance criteria are heuristically defined and may not accurately quantify a parameter's contribution to the learning process: (1) magnitude-based criteria measure the importance of parameters by their absolute values. However, even parameters with small magnitude can significantly impact model performance due to the complex compositional structure of neural language models; (2) Sensitivity-based criteria estimate the importance of a parameter by the absolute change in training error if only that parameter is pruned and others remain unchanged. While taking the learning objective into account, such an approach still determines the importance of each parameter individually without considering the effect of coinstantaneous parameter updates on model performance. 

In this paper, we rethink the design of the importance criterion for model pruning from an optimization perspective. We begin by analyzing the temporal variation of any given learning objective based on a single-step gradient descent update under the iterative pruning setting. We show that finding the optimal pruning decision can be framed as solving an equality-constrained 0-1 Integer Linear Programming~(ILP) problem, where the constraint is defined by the specified sparsity. The resulting problem is a particular case of a general 0-1 Knapsack problem in which the weight for each item is the same. The solution to this problem naturally leads to a principled importance criterion which we use to rank all model parameters and derive the optimal stepwise pruning decision.

When a high sparsity~(e.g., 80\%$\sim$90\%) is pursued, the limited capacity often renders the pruned model fails to retain satisfactory performance with conventional fine-tuning. To further improve the model's generalization ability, we propose a self-regularization scheme, where the model prediction is regularized by the latest best-performing model checkpoint during pruning. We show that such a scheme eases model learning with decreasing capacity and effectively yields a tighter upper bound of expected generalization error than learning from training data alone.

To validate the effectiveness of our approach, dubbed PINS~(\underline{P}runing with principled \underline{I}mportance a\underline{N}d \underline{S}elf-regularization), we conducted extensive experiments with various pre-trained language models on a wide variety of tasks, including natural language understanding on GLUE~\cite{glue}), question answering on SQuAD~\cite{squad}, named entity recognition on CoNLL 2003~\cite{conll2003}, and data-to-text generation on WebNLG~\cite{webnlg}, DART~\cite{dart}, and E2E~\cite{e2e}. Experimental results show that PINS provides more accurate models at different sparsity levels. Detailed analysis shed further light on some intriguing properties of models pruned by PINS. By exploiting the resulting high sparsity, we show that the storage/inference can be reduced/accelerated by 8.9x and 2.7x using CSR format and a sparsity-aware inference runtime~\cite{deepsparse} on consumer-level CPUs~\footnote{Code available at \url{https://github.com/DRSY/PINS}}.

In summary, our contributions are:

\begin{itemize}
	
	\item We establish the equivalence between the optimal pruning decision and the solution to an equality-constrained 0-1 Integer Linear Programming problem. The solution to this problem leads to a principled importance criterion that can be used to rank parameters during iterative pruning.

\item We propose a simple yet effective self-regularization scheme to enhance the model's generalization capability, especially under a high-sparsity regime.

\item Comprehensive experiments and analyses confirm the effectiveness of our approach at various sparsity levels. 
	
\end{itemize}

	\section{Background and Related Work}
In this section, we review the necessary background on Transformer-based pre-trained language models and popular importance criteria for iterative pruning.
\subsection{Transformer-based Pre-trained Language Models}
Most existing pre-trained neural language models~\cite{gpt2,bert,minilm,electra} are based on the Transformer~\cite{transformer} architecture, which consists of several identical blocks of self-attention and feedforward network. After pre-training on a massive amount of unlabeled general-domain corpus in a self-supervised learning manner, these models exhibit superior performance on various downstream tasks via fine-tuning. However, good generalization performance comes at the cost of a vast amount of parameters. For example, the base version of BERT has 110M parameters and leads to more than 400MB of disk storage. Therefore, how to effectively reduce model size while preserving as much task accuracy as possible remains a challenging research problem.
\subsection{Iterative Pruning}
Pruning methods can be divided into two categories: one-shot pruning~\cite{oneshot1,oneshot2} and iterative pruning~\cite{l0,movement,platon}. One-shot pruning removes parameters of low importance after training. It is efficient but ignores the complicated training dynamics when applied to modern large neural language models. On the contrary, iterative pruning performs training and pruning simultaneously. Therefore, the resulting sparsity pattern is aware of the complex dynamics of parameters through the course of training and delivers considerable improvement compared to one-shot pruning.

Let $\bm{\theta}^{(t)}=\{\theta_1^{(t)}\,\theta_2^{(t)},...,\theta_d^{(t)}\}$ denote the $d$-dimensional model parameters at $t$-th training iteration, the typical updating rule of iterative pruning can be formulated as:
\begin{align}
	\hat{\bm{\theta}}^{(t+1)}&=\bm{\theta}^{(t)}-\eta^{(t)}\nabla_{\bm{\theta}} \mathcal{L}(\bm{\theta}^{(t)}) \\
	\bm{\theta}^{(t+1)} &= \hat{\bm{\theta}}^{(t+1)} \odot \bm{M}^{(t)}
	\label{eq:updaterule}
\end{align}
where $\eta^{(t)}$ is the learning rate at time step $t$ and $\mathcal{L}$ is the learning objective. The temporarily updated $\hat{\bm{\theta}}^{(t+1)}$ is further pruned by the binary mask $\bm{M}^{(t)}$$\in$ $\{0,1\}^{d}$, which is computed based on a given importance criterion $\bm{S}^{(t)}$:
\begin{align}
	\bm{M}^{(t)}_i=
	\begin{cases} 
		1, & \text{if }~\bm{S}^{(t)}_i\text{is in the top-}r^{(t)}\text{of }\bm{S}^{(t)}\\
		0,  & \text{otherwise}  
	\end{cases}
\label{eq:mask}
\end{align}
where $r^{(t)}$$\leq d$ indicates the number of remaining parameters at time step $t$ according to a given sparsity scheduler.
\subsection{Importance Criteria for Model Pruning}
Popular importance criteria for model pruning include parameters' magnitude and sensitivity.
\paragraph{Magnitude} is a simple yet effective importance criterion that is widely used for model pruning. It estimates the importance of each parameter as its absolute value, i.e., $\bm{S}^{(t)}_i=|\bm{\theta}^{(t)}_i|$. Despite its simplicity, the magnitude cannot accurately gauge the importance of parameters because even parameters with small magnitude can have a large impact on the model prediction due to the complex compositional structure of PLMs.

\paragraph{Sensitivity} is another useful importance criterion. It estimates the importance of each parameter as the absolute change of the learning objective if the parameter is pruned, i.e., set to zero. The mathematical formulation of the sensitivity of $i$-th parameter is given by:
\begin{align}
	\bm{S}^{(t)}_i& =|\mathcal{L}(\bm{\theta}^{(t)}_{-i})-\mathcal{L}(\bm{\theta}^{(t)})| \\
	&\approx |\bm{g}_i^{(t)}\bm{\theta}^{(t)}_i|
\end{align}
where $\bm{\theta}^{(t)}_{-i}$ is identical to $\bm{\theta}^{(t)}$ except that the $i$-th entry is set to zero and $g_i^{(t)}$ is the gradient of $i$-th entry. Though taking the training dynamics into account, sensitivity still estimates the importance of each parameter individually without considering the effect of holistic parameter update.

	\section{Methodology}
Instead of heuristically defining the importance criterion as in prior pruning methods, we take a step back and rethink the design of the importance criterion for model pruning from an optimization perspective. From our analysis, we draw an equivalence between finding the optimal stepwise pruning decision and solving an equality-constrained 0-1 Integer Linear Programming problem. We further show that the optimal solution to this problem leads to a new importance criterion for model pruning. Moreover, we propose a simple yet effective self-regularization scheme to facilitate the generalization ability of the sparse model. We elucidate our analysis in \secref{sec:analysis} and describe our self-regularization scheme in \secref{sec:sr}.
% \subsection{Analysis}
\subsection{Rethinking Importance Criterion from the Optimization Perspective}
\label{sec:analysis}
Without loss of generality, we denote $\mathcal{L}$ as the learning objective when adapting a pre-trained language model $f$ with parameter $\bm{\theta}$ to a downstream task. At $t$-th training iteration, we denote the current model parameters as $\bm{\theta}^{(t)}$ and the evaluated learning objective as $\mathcal{L}(\bm{\theta}^{(t)})$.

The temporal variation of the learning objective $\mathcal{L}(\bm{\theta}^{(t)})$ at time step $t$ is given by the second-order Taylor series expansion:
\begin{align}
	\Delta\mathcal{L}^{(t)}&=\mathcal{L}(\bm{\theta}^{(t)}+\Delta \bm{\theta}^{(t)})-\mathcal{L}(\bm{\theta}^{(t)}) \\ \nonumber
	&= \nabla_{\bm{\theta}}\mathcal{L}(\bm{\theta}^{(t)})^{\top}\Delta \bm{\theta}^{(t)}+ \\
	& \frac{1}{2}\Delta \bm{\theta}^{(t)^{\top}}\bm{H}^{(t)}\Delta \bm{\theta}^{(t)}+o(|\Delta\bm{\theta}^{(t)}|^2)
	\label{eq:7}
\end{align}
where $\bm{H}^{(t)}$ is the Hessian matrix at step $t$. 
It is known that the largest eigenvalue $\lambda_{max}$ of Hessian matrices in 
a PLM is typically small~\cite{eigen}, i.e., $\Delta \bm{\theta}^{(t)^{\top}}\bm{H}^{(t)}\Delta \bm{\theta}^{(t)}\leq\lambda_{max}|\Delta \bm{\theta}^{(t)}|_2^2\approx 0$. Thus, we ignore the second-order term as well as the infinitesimal of higher order in \eqnref{eq:7}:
\begin{align} \nonumber
	\Delta\mathcal{L}^{(t)}&=\nabla_{\bm{\theta}}\mathcal{L}(\bm{\theta}^{(t)})^{\top}\Delta \bm{\theta}^{(t)} \\
	&=\sum_{i=1}^{d}\bm{g}_i^{(t)}\cdot \Delta \bm{\theta}^{(t)} _i
	\label{eq:8}
\end{align}
Under the iterative pruning setting, the actual temporal variation $\Delta\bm{\theta}^{(t)}_i$ of $i$-th parameter depends on whether it is allowed to be updated or forced to zeroed out. Formally, we use a binary variable $\bm{x}_i^{(t)}$ to indicate the pruning decision of $i$-th parameter at time step $t$, i.e., $\bm{x}_i^{(t)}=1$ means $\bm{\theta}^{(t)}_i$ is updated and $\bm{x}_i^{(t)}=0$ means $\bm{\theta}^{(t)}_i$  is pruned. 
%\KZ{Earlier we used $M_i^t$ as the binary mask, now we are using $x$. A bit confusing. I also find the notation in this subsection to be too complicated.}
The temporal variation in \eqnref{eq:8} can now be rewritten as:
\begin{align}
	\Delta\mathcal{L}^{(t)}=\sum_{i=1}^{d}\bm{g}_i^{(t)}(\bm{x}_i^{(t)}\Delta\hat{\bm{\theta}}_i^{(t)}+(1-\bm{x}_i^{(t)})(-\bm{\theta}_i^{(t)}))
\end{align}
where $\Delta\hat{\bm{\theta}}_i^{(t)}=-\eta^{(t)}\bm{g}_i^{(t)}$ is the gradient descent update.  Finding the optimal pruning decision that leads to the smallest $\Delta\mathcal{L}^{(t)}$ is now converted to an equality-constrained 0-1 integer linear programming~(ILP) problem of variables $\bm{x}^{(t)}$:
\begin{align}
%	\label{eq:ks}
\nonumber
	\tilde{\bm{x}}^{(t)}&=\underset{\bm{x}^{(t)}}{\arg\min} ~\Delta\mathcal{L}^{(t)} \\
	\text{s.t.~~~~~}\sum_{i=1}^d &\bm{x}_i^{(t)}=r^{(t)}, \bm{x}_i^{(t)}\in\{0,1\} 
	\label{eq:10}
\end{align}
where $r^{(t)}$ is the number of remaining parameters at step $t$ according to the pre-defined sparsity scheduler. 
%\KZ{I find this scheduler to be a bit 
%interesting. The overall goal should be to hit a predefined sparcity, such as 80\%, but how to reach there in each iteration may not be steady speed. So I 
%wonder how this scheduler works. It seems that every iteration theres a sparsity
%budget which is the $v^{(t)}$, but who determines this?}
If we consider each parameter $\bm{\theta}^{(t)}_i$ as an item and $r^{(t)}$ as the total capacity,  the problem that \eqnref{eq:10} defines can be treated as a special case of 0-1 Knapsack problem where the weight for each item is one and the value for each item is given by:
%Instead of relying on an external specialized solver~\cite{ilp1}, we prove that the optimal solution $\tilde{\bm{x}}^{(t)}$ can be efficiently derived based on the following new importance criteria:
\begin{align}
	\bm{S}_i^{(t)}=-\bm{g}_i^{(t)}\Delta\hat{\bm{\theta}}_i^{(t)}-\bm{g}_i^{(t)}\bm{\theta}_i^{(t)}
	\label{eq:score}
\end{align}
Contrary to the general 0-1 Knapsack problem which is known to be NP-complete, fortunately, the equal-weight 0-1 Knapsack is a P problem.  Its optimal solution can be obtained by sorting items in descending order according to their values and selecting the top-$r^{(t)}$ ones:
\begin{align}
	\tilde{\bm{x}}^{(t)}_i=
	\begin{cases} 
		1, & \text{if }~\bm{S}^{(t)}_i\text{is in the top-}r^{(t)}\text{of }\bm{S}^{(t)}\\
		0,  & \text{otherwise}  
	\end{cases}
	\label{eq:problem}
\end{align}
%\KZ{This equation seems to be very similar to Eq (3)?}

%Putting it in the context of iterative pruning, we can interpret the value defined in \eqnref{eq:score} as a principled new importance criterion and use it to rank all model parameters.
Putting it in the context of iterative pruning, our analysis theoretically reveals the validity of: (1) selecting parameters based on the ranking of certain importance criterion; (2) using \eqnref{eq:score} as a  principled new importance criterion.

\subsection{Self-regularization}
\label{sec:srr}
In vanilla fine-tuning, the learning objective $\mathcal{L}$ is defined as the training error $\mathcal{L}_{er}$~(a.k.a empirical risk in statistical learning) over the empirical data distribution.  However, minimizing such training error does not translate to good generalization. Moreover, as iterative pruning proceeds, the number of non-zero parameters in the model monotonically decreases. The reduced model capacity increases the learning difficulty~\cite{difficulty1,difficulty2} and usually leads to degenerated generalization performance of the sparsified model~\cite{movement}. 

Confronting the above challenges, we propose an effective self-regularization scheme tailored to improving the model's generalization ability during iterative pruning. Concretely, besides learning from the hard label of training data, the output of the current model with parameter $\bm{\theta}^{(t)}$ is also regularized by the output of the latest best-performing model checkpoint with parameter $\bm{\theta}^{(t_l)}$, where $t_l\leq t$ denotes the time step at which the latest checkpoint was saved. The learning objective of self-regularization is defined as:
\begin{align}
	\mathcal{L}_{sr}=\mathcal{D}(y_{\bm{\theta}^{(t)}},y_{\bm{\theta}^{(t_l)}})
\end{align}
where $\mathcal{D}$ can be any divergence metric, e.g., KL-divergence for classification tasks. $\mathcal{L}_{sr}$ is then integrated with the original learning objective, i.e., $\mathcal{L}=\mathcal{L}_{er}+\mathcal{L}_{sr}$.
\paragraph{Why does self-regularization work?}Our self-regularization is similar to teacher-student knowledge distillation in the sense that the model output is regularized by the output of another model. However, the most critical difference is that the ``teacher'' in self-regularization is instantiated by checkpoint with increasing sparsity, such that the capacity gap between ``teacher'' and ``student'' is dynamically adjusted. We theoretically justify the effectiveness of self-regularization as follows:
% \KZ{I don't quite understand the 
%``expected generalization error'' $R$. Is there such a thing? How do you 
%quantify that? We gotta be careful when we are using these special notations>}
\newtheorem{theorem}{Theorem}
\begin{theorem}
	Let $t_i$ and $t_{j}$ where $t_{i}\geq t_{j}$ denote the time steps at which two different checkpoints are saved; Let $R(f_{\bm{\theta}^{(t\leftarrow t_i)}})$ and $R(f_{\bm{\theta}^{(t\leftarrow t_j)}})$ denote the expected generalization error of models learned from $f_{\bm{\theta}^{(t_i)}}$ and $f_{\bm{\theta}^{(t_j)}}$; Let n denotes the size of training data; $|\cdot|_{\text{C}}$ denotes a capacity measure of function class $\mathcal{F}_{\bm{\theta}}$. Based on previous expositions on VC theory~\cite{vc}, we have the following asymptotic generalization bounds hold:
	\begin{align}\nonumber
		R(f_{\bm{\theta}^{(t\leftarrow t_i)}})\leq \underbrace{O(\frac{|\mathcal{F}_{\bm{\theta}^{(t)}}|_{\text{C}}}{n^{\alpha_{i}}})+\underset{ \mathcal{F}_{\bm{\theta}^{(t\leftarrow t_i)}}}{\inf}R(f_{\bm{\theta}^{(t)}})}_{bound(f_{\bm{\theta}^{(t\leftarrow t_i)}})} \\
		R(f_{\bm{\theta}^{(t\leftarrow t_j)}})\leq \underbrace{O(\frac{|\mathcal{F}_{\bm{\theta}^{(t)}}|_{\text{C}}}{n^{\alpha_{j}}})+\underset{ \mathcal{F}_{\bm{\theta}^{(t\leftarrow t_j)}}}{\inf}R(f_{\bm{\theta}^{(t)}})}_{bound(f_{\bm{\theta}^{(t\leftarrow t_j)}})}  \nonumber
	\end{align}
Because $\bm{\theta}^{(t_i)}$ is a later checkpoint with higher sparsity than $\bm{\theta}^{(t_j)}$, we have the learning speed $1\geq \alpha_{i}\geq \alpha_{j}\geq \frac{1}{2}$, then the following inequality holds with high probability:
\begin{align}\nonumber
	bound(f_{\bm{\theta}^{(t\leftarrow t_i)}}) \leq bound(f_{\bm{\theta}^{(t\leftarrow t_j)}})
\end{align}
\end{theorem}
In summary, self-regularization works by enabling a tighter generalization bound compared to learning from training data alone or a static dense teacher as in knowledge distillation. Please refer to Appendix \ref{sec:B} for detailed derivation.
\label{sec:sr}

\subsection{The Algorithm}
Here we formally summarize our algorithm PINS~(\underline{P}runing with principled \underline{I}mportance a\underline{N}d \underline{S}elf-regularization) in Algorithm \ref{alg:alg}:
\begin{algorithm}[h]
	\caption{PINS} %算法的名字
	\hspace*{0.02in} {\bf Input:} %算法的输入， \hspace*{0.02in}用来控制位置，同时利用 \\ 进行换行
	Training set $\mathcal{D}_{tr}=\{(x_i, y_i)\}_{i=1}^N$; Validation set $\mathcal{D}_{val}$; pre-trained parameters $\bm{\theta}$;  maximum training steps $T$; evaluation interval $t_{eval}$. \\
	\textbf{Initialize:} $\bm{\theta}^{(0)}\leftarrow \bm{\theta}$, $t_{l}\leftarrow 0$, best validation accuracy $\text{acc}_{t_l}\leftarrow -\text{INF}$.
	\begin{algorithmic}[1]
		\For{$t=0$ to $T-1$}
				 \State Sample a mini-batch  $(\bm{x}, \bm{y})$ from $D_{tr}$
				 \State Compute current model's output $\bm{y}_{\bm{\theta}^{(t)}}$
				 \State Compute latest best-performing checkpoint's output $\bm{y}_{\bm{\theta}^{(t_l)}}$
				 \State Compute $\mathcal{L}$ based on $\bm{y}_{\bm{\theta}^{(t)}}$, $\bm{y}_{\bm{\theta}^{(t_l)}}$ and $\bm{y}$
				 \State Compute $\bm{S}^{(t)}$ via \eqnref{eq:score}
				 \State Compute $\bm{\theta}^{(t+1)}$ via \eqnref{eq:updaterule} and \eqnref{eq:mask}
				 \If{$t \% t_{eval}=0$ and acc$_{t}$>acc$_{t_l}$}
				  \State acc$_{t_l}\leftarrow\text{acc}_{t}$, $\bm{\theta}^{(t_l)}\leftarrow\bm{\theta}^{(t)}$
				 \EndIf
		\EndFor

	\end{algorithmic}
	\hspace*{0.02in} {\bf Output:} %算法的结果输出
the pruned parameters $\bm{\theta}^{(T)}$.
	\label{alg:alg}
\end{algorithm}
%\subsection{PINS: Pruning with Principled Weight Importance and Self-regularization}

	\section{Experiments}
In this section,  We compare PINS with state-of-the-art pruning algorithms and perform detailed analysis  to understand the effectiveness of PINS.
\subsection{Setup}
\subsubsection{Tasks}
We conduct experiments on a comprehensive spectrum of tasks following standard data splits. \\
\textbf{Natural Language Understanding. } We opt for tasks from the GLUE~\cite{glue} benchmark, including linguistic acceptability~(CoLA), natural language inference~(RTE, QNLI, MNLI), paraphrase~(MRPC, QQP), sentiment analysis~(SST-2) and textual similarity~(STS-B). Because the official test set of GLUE is hidden, we randomly split a small portion of training set as validation set and treat the original validation set as test set. \\
\textbf{Question Answering. } We use SQuAD v1.1~\cite{squad} as a representative dataset for extractive question answering following previous work~\cite{platon}. \\
\textbf{Named Entity Recognition. } We also examine our approach on CoNLL 2003~\cite{conll2003} for token-level named entity recognition task. \\
\textbf{Data-to-Text Generation.} Besides language understanding tasks, we also extend our evaluation to data-to-text generation on three datasets: E2E~\cite{e2e}, DART~\cite{dart}, and WebNLG~\cite{webnlg}, which involves generating a piece of fluent text from  a set of structured relational triples. \\

\subsubsection{Baselines}
\textbf{Magnitude-based.} Iterative magnitude pruning~(IMP)~\cite{gupta} is the state-of-the-art magnitude-based approach.

\noindent
\textbf{Sensitivity-based.} $l_0$-regularization~\cite{l0} trains masking variables via re-parametrization trick with $l_0$ penalty; SMvP~\cite{movement} uses accumulated sensitivity as importance metric; PST~\cite{pst} proposed a hybrid importance criterion combining both magnitude and sensitivity; PLATON~\cite{platon} uses a modified variant of sensitivity by exponential moving average and uncertainty re-weighting.

%\subsubsection{Models}

\begin{table*}[t]
	\centering
	\normalsize
	\small
	\begin{tabular}{c|l|cccccccc|c}
		\toprule
		\multirow{2}{*}{Sparsity} & \multicolumn{1}{c|}{\multirow{2}{*}{Method}} & \multicolumn{1}{c}{\multirow{2}{*}{\begin{tabular}[c]{@{}c@{}}\textbf{RTE}\\ Acc\end{tabular}}} & \multicolumn{1}{c}{\multirow{2}{*}{\begin{tabular}[c]{@{}c@{}}\textbf{MRPC}\\ F1\end{tabular}}} & \multicolumn{1}{c}{\multirow{2}{*}{\begin{tabular}[c]{@{}c@{}}\textbf{STS-B}\\ Pearson\end{tabular}}} & \multicolumn{1}{c}{\multirow{2}{*}{\begin{tabular}[c]{@{}c@{}}\textbf{CoLA}\\ Mcc\end{tabular}}} & \multicolumn{1}{c}{\multirow{2}{*}{\begin{tabular}[c]{@{}c@{}}\textbf{SST-2}\\ Acc\end{tabular}}} & \multicolumn{1}{c}{\multirow{2}{*}{\begin{tabular}[c]{@{}c@{}}\textbf{QNLI}\\ Acc\end{tabular}}} & \multicolumn{1}{c}{\multirow{2}{*}{\begin{tabular}[c]{@{}c@{}}\textbf{MNLI}\\ Acc\end{tabular}}} & \multicolumn{1}{c|}{\multirow{2}{*}{\begin{tabular}[c]{@{}c@{}}\textbf{QQP}\\ Acc\end{tabular}}} & \multicolumn{1}{c}{\multirow{2}{*}{Avg.}}\\
		& \multicolumn{1}{c|}{}                        & \multicolumn{1}{c}{}                                                                   & \multicolumn{1}{c}{}                                                                   & \multicolumn{1}{c}{}                                                                    & \multicolumn{1}{c}{}                                                                     & \multicolumn{1}{c}{}                                                                    & \multicolumn{1}{c}{}                                                                    & \multicolumn{1}{c}{}             & \multicolumn{1}{c|}{}                &\multicolumn{1}{c}{}                                                  \\
		\midrule
		\multicolumn{1}{c|}{0\%}   & Fine-tune$\dagger$                                   & 69.3                                                                                   &
		90.3                                                                            & 90.2       & 58.3                                                                                    & 92.4                                                                                     & 91.3                                                                                    & 84.0                                                                                    & 91.5                                 &  83.4                                                \\
		\midrule
		\multirow{6}{*}{80\%}     &IMP$\dagger$                                                 &65.7                                                                                        &86.2                                                                               & 86.8        & 42.5                                                                                        &84.3                                                                                 &89.2                                                                                     &82.2                                                                                        &  86.0                                                    &    77.9                              \\
		&$l_0$-regularization$\dagger$                                                 & 63.2                                                                                       & 80.2                                                                                 & 82.8      &0.0                                                                                         &   85.0                                                                                       &       85.0                                                                                  &    80.8                                                                                     &  88.5                                                          &  70.7                          \\
		&SMvP$\dagger$                                                 &62.8                                                                                        &   86.7                                                                    & 87.8                 & 48.5                                                                                        &89.0                                                                                          &  88.3                                                                                       &   81.9                                                                                      &    90.6                                                                 & 79.5                  \\
		&PST                                             &          63.0                                                                              &       87.4   & 88.0                                                                              &      44.6                                                                                   &  89.3                                                                                        &    88.3                                                                                     &    79.3                                                                                     &  88.9                                                                    &  78.6                \\
		&PLATON$\dagger$                                             &   68.6                                                                                     & 89.8                                                                                    &89.0   &     54.5                                                                                    & 91.2                                                                                         &   90.1                                                                                      &    83.3                                                                                     &    90.7                                                                 &82.2                   \\
		&PINS~(ours)                                             &\textbf{72.7}                                                                                       &\textbf{90.9}                                                                                      &\textbf{89.2}   & \textbf{57.1}                                                                                      & \textbf{91.9}                                                                                       &\textbf{91.2}                                                                                         &  \textbf{83.9}                                                                                      &  \textbf{90.9}                                                                    & \textbf{83.5}               \\
		\midrule
		\multirow{6}{*}{90\%}     &IMP$\dagger$                                                 &57.4                                                                                       &80.3                                                                           &83.4             & 18.3                                                                                        &  80.7                                                                                        & 86.6                                                                                        &78.9                                                                                         &78.8                                                                     &  70.5                 \\
		&$l_0$-regularizatio$\dagger$                                                 &59.9                                                                                        &  79.5                                                                               &  82.7     & 0.0                                                                                        &    82.5                                                                                      &   82.8                                                                                      &    78.4                                                                                     &   87.6                                                                           &69.1          \\
		&SMvP$\dagger$                                                 &58.8                                                                                        &   85.9                                                                                &86.5     &  0.0                                                                                       &   87.4                                                                                       &  86.6                                                                                       &  80.9                                                                                       &    90.2                                                                        &  72.1          \\
		&PST$\ddagger$                                             &62.8                                                                                        &  85.6                                                                             &81.7         & 42.5                                                                                        & 88.7                                                                                         &   86.0                                                                                      &    76.7                                                                                     & 83.9                                                                              & 76.0        \\
		& PLATON$\dagger$                                                &  65.3                                                                                      & 88.8                                                                               &87.4        &44.3                                                                                         &  90.5                                                                                        &88.9                                                                                         &  81.8                                                                                       &  90.2                                                                              &   79.6     \\
		&PINS~(ours)                                             &  \textbf{68.5}                                                                                      & \textbf{90.1}                                                                              &\textbf{87.9}         & \textbf{49.8}                                                                                        &  \textbf{91.0}                                                                                        &  \textbf{89.5}                                                                                       &   \textbf{82.7}                                                                                      &  \textbf{90.6}                                   & \textbf{81.3}            \\
		\bottomrule                                     
	\end{tabular}
	\caption{Results with BERT$_{\text{base}}$ on the GLUE development set. For MNLI, the results are averaged on MNLI-m and MNLI-mm. $\dagger$ indicates the results are directly quoted from \citet{platon} while $\ddagger$ indicates the results are reported by \citet{pst}. }
	\label{table:glue}
\end{table*}

\subsubsection{Implementation Details}
We mainly conduct experiments on the pre-trained BERT$_{\text{base}}$~\cite{bert} as a pruning target for all tasks except data-to-text generation. We defer the pruning results of MiniLM$_{\text{12L-384H}}$~\cite{minilm} and Electra$_{\text{base}}$~\cite{electra} to Appendix \ref{sec:A}. For data-to-text generation, we adopt the pre-trained GPT-2~\cite{gpt2} following 
a prior study~\cite{pst}.

During pruning, we employ the cubic sparsity scheduler~\cite{movement,platon} to gradually increase the sparsity level from 0 to the specified target sparsity. To avoid tremendous computation cost brought by hyper-parameter tuning, we only search the batch size from $\{16,32\}$ and fix the learning rate as 3e-5 for all experiments on GLUE and CoNLL. For SQuAD v1.1, we fix the batch size as 16 and the learning rate as 3e-5 following \citet{platon}. We adopt AdamW~\cite{adamw} as the default optimizer.
To reduce the variance induced by mini-batch sampling, we adopt a smoothing technique similar to PLATON. We run each experiment five times with different random seeds and report the average results~(significance tests with $p$-value~<~0.05 are conducted  for  all performance gains).

\subsection{Main Results}

\subsubsection{Comparison with Baselines}

\begin{table}[t]
	\centering
	\small
	\begin{tabular}{l|cccc}
		\toprule
		Sparsity   & 80\% & 70\% & 60\% & 50\% \\
		\midrule
		Fine-tune$\dagger$  & \multicolumn{4}{c}{88.1}  \\
		\midrule
		IMP$\dagger$        & 82.9 & 86.5 & 86.7 & 87.0 \\
		$l_0$-regularization$\dagger$      & 81.9 & 82.8 & 83.9 & 84.6 \\
		SMvP$\dagger$       & --& 84.6 & -- & 85.8 \\
		PLATON$\dagger$     & 86.1 & 86.7 & 86.9 & 87.2 \\
		\midrule
		PINS~(ours) & \textbf{86.4} & \textbf{86.9} & \textbf{87.4} & \textbf{88.0} \\
		\bottomrule
	\end{tabular}
	\caption{Results with BERT$_{\text{base}}$ on SQuAD v1.1. $\dagger$ indicates numbers reported from \citet{platon}. F1 score is reported as evaluation metric.}
	\label{table:qa}
\end{table}
\paragraph{Natural language understanding} We present the experimental results on GLUE at high sparsity, i.e., 80\% and 90\% in  \tabref{table:glue}. Among all baselines, sensitivity-based methods generally achieve better results than magnitude-based IMP, which implies the importance of training dynamics when designing pruning criteria. We can see that PINS delivers more accurate sparsified models on all datasets at both sparsity levels. The advantage of PINS is more evident on small datasets. For example, PINS outperforms the previous best-performing baseline~(PLATON) by 4.1 and 2.6 points on RTE and CoLA at 80\% sparsity, where there are only a few thousand training data. Under extremely high sparsity, i.e., 90\%, PINS is still able to retain 97.5\% overall performance of fine-tuning, outperforming 95.4\% of the previous best method PLATON. Notably, PINS even surpasses fine-tuning on RTE and MRPC at 80\% sparsity. This can be attributed to the fact that PLMs are heavily over-parameterized and PINS can effectively identify parameters crucial to the task to realize low bias and low variance simultaneously.

% Please add the following required packages to your document preamble:
% \usepackage{multirow}
\begin{table}[t]
	\centering
	\small
	\begin{tabular}{c|lccc}
		\toprule
		Sparsity              & Method     & P & R & F1   \\
		\midrule
		0\%                   & Fine-tune  & 93.5      & 94.6   & 94.0 \\
		\midrule
		\multirow{3}{*}{70\%} & IMP        & 90.7      & 91.8   & 91.2 \\
		& SMvP       & 92.9      & 94.1   & 93.5 \\
		& PINS(ours) & \textbf{93.5}      & \textbf{94.3}   & \textbf{93.9} \\
		\midrule
		\multirow{3}{*}{80\%} & IMP        & 84.4      & 87.3   & 85.8 \\
		& SMvP       & 92.1      & 93.1   & 92.6 \\
		& PINS(ours) & \textbf{92.8}      & \textbf{93.8}   & \textbf{93.3} \\
		\bottomrule
	\end{tabular}
	\caption{Results with BERT$_{\text{base}}$ on CoNLL 2003. P and R stands for Precision and Recall respectively.}
	\label{table:ner}
\end{table}

% Please add the following required packages to your document preamble:
% \usepackage{multirow}
\begin{table*}[t]
	\centering
	\small
	\begin{tabular}{c|l|ccc|cc|cc}
		\toprule
		\multirow{2}{*}{Sparsity} & \multicolumn{1}{c|}{\multirow{2}{*}{Method}} & \multicolumn{3}{c|}{\textbf{E2E}}                                                             & \multicolumn{2}{c|}{\textbf{DART}}                                                        & \multicolumn{2}{c}{\textbf{WebNLG}}                                                      \\
		& \multicolumn{1}{c|}{}                        & \multicolumn{1}{l}{BLEU} & \multicolumn{1}{l}{ROUGE-L} & \multicolumn{1}{l|}{METEOR} & \multicolumn{1}{l}{BLEU} & \multicolumn{1}{l|}{BLEURT} & \multicolumn{1}{l}{BLEU} & \multicolumn{1}{l}{BLEURT} \\
		\midrule
		0\%                       & Fine-tune                                   & 69.4                     & 71.1                        & 46.2                       & 46.6                                        & 0.30                       & 46.9                                       & 0.23                       \\
		\midrule
		\multirow{3}{*}{80\%}     & IMP                                         & 69.3                     & 71.0                        & 45.8                       & 44.9                                   & 0.22                       & 39.9                                      & 0.00                       \\
		& PST                                         & 69.4                     & 70.8                        & 45.9                       & 44.1                                  & 0.22                       & 44.3                                   & 0.16                       \\
		& PINS~(ours)                                        & \textbf{69.6}                     & \textbf{71.8}                        & \textbf{46.6}                       & \textbf{46.2}                                       & \textbf{0.29}                       & \textbf{45.5}                                       & \textbf{0.18}                       \\
		\bottomrule
	\end{tabular}
	\caption{Results with GPT-2 on data-to-text generation datasets.  The higher the BLEU, ROUGE-L, METEOR, and BLEURT scores are, the better the performance.}
	\label{table:dtt}
\end{table*}
\begin{table*}[t]
	\centering
	\small
	\begin{tabular}{c|l|cccccccc|c}
		\toprule
		\multirow{2}{*}{Sparsity} & \multicolumn{1}{c|}{\multirow{2}{*}{Method}} & \multicolumn{1}{c}{\multirow{2}{*}{\begin{tabular}[c]{@{}c@{}}\textbf{RTE}\\ Acc\end{tabular}}} & \multicolumn{1}{c}{\multirow{2}{*}{\begin{tabular}[c]{@{}c@{}}\textbf{MRPC}\\ F1\end{tabular}}} & \multicolumn{1}{c}{\multirow{2}{*}{\begin{tabular}[c]{@{}c@{}}\textbf{STS-B}\\ Pearson\end{tabular}}} & \multicolumn{1}{c}{\multirow{2}{*}{\begin{tabular}[c]{@{}c@{}}\textbf{CoLA}\\ Mcc\end{tabular}}} & \multicolumn{1}{c}{\multirow{2}{*}{\begin{tabular}[c]{@{}c@{}}\textbf{SST-2}\\ Acc\end{tabular}}} & \multicolumn{1}{c}{\multirow{2}{*}{\begin{tabular}[c]{@{}c@{}}\textbf{QNLI}\\ Acc\end{tabular}}} & \multicolumn{1}{c}{\multirow{2}{*}{\begin{tabular}[c]{@{}c@{}}\textbf{MNLI}\\ Acc\end{tabular}}} & \multicolumn{1}{c|}{\multirow{2}{*}{\begin{tabular}[c]{@{}c@{}}\textbf{QQP}\\ Acc\end{tabular}}} & \multicolumn{1}{c}{\multirow{2}{*}{Avg.}}\\
		& \multicolumn{1}{c|}{}                        & \multicolumn{1}{c}{}                                                                   & \multicolumn{1}{c}{}                                                                   & \multicolumn{1}{c}{}                                                                    & \multicolumn{1}{c}{}                                                                     & \multicolumn{1}{c}{}                                                                    & \multicolumn{1}{c}{}                                                                    & \multicolumn{1}{c}{}             & \multicolumn{1}{c|}{}                        &\multicolumn{1}{c}{}                                            \\
		\midrule
		\multicolumn{1}{c|}{0\%}   & Fine-tune                                   & 69.3                                                                                   &
		90.3                                                                        &90.2           & 58.3                                                                                    & 92.4                                                                                     & 91.3                                                                                    & 84.0                                                                                    & 91.5                                 &  83.4                                                \\
		\midrule
		\multirow{1}{*}{50\%}   &PINS                                             & 70.8                                                                                     &  91.4                                                                                  &89.7   &    60.6                                                                          &    92.9                                                                                 &   91.8                                                                              &  85.1                                                                                    &      91.3                                                             & 84.2              \\
		\multirow{1}{*}{30\%} 
		&PINS                                            &    71.7                                                                                   & 91.2                                                                             &89.8    &60.4                                                                                   &  93.3                                                                                      &  92.0                                                                                    &      85.1                                                                               &  91.5                                &  84.4       \\
		\bottomrule                                     
	\end{tabular}
	\caption{Results with BRET$_{\text{base}}$ on the GLUE development set under medium-to-low sparsity regime. Numbers are the mean of five trials with different random seeds. PINS outperforms fine-tuning at medium-to-low sparsity.}
	\label{table:mtl}
\end{table*}
%\KZ{Can you give an explanation for IMP's good results?} 
\paragraph{Question answering} \tabref{table:qa} summarizes the pruning results on SQuAD v1.1. Interestingly, IMP outperforms all sensitivity-based methods except for PLATON at all considered sparsity levels, in contrast to the observations on GLUE. Our method, however,  consistently yields the best performance at all sparsity settings.
\paragraph{Named entity recognition} \tabref{table:ner} demonstrates the pruning results on CoNLL 2003 dataset for named entity recognition. At 70\% sparsity, our method almost matches the performance of fine-tuning, outperforming baselines on all evaluation metrics. The gain of PINS is more prominent when further increasing sparsity.

\paragraph{Data-to-text generation} 
%We also conduct experiments on data-to-text generation tasks on three datasets to examine the effectiveness of our method beyond NLU tasks. 
\tabref{table:dtt} shows the pruning results on E2E, DART and WebNLG at 80\% sparsity. PINS achieves the best performance on all three datasets in all evaluation metrics. In particular, PINS delivers performance even better than fine-tuning on the E2E dataset by 0.7 ROUGE-L and 0.4 METEOR scores, respectively. We posit that this is due to the relative easiness of E2E compared to the other two datasets.

\subsubsection{Results at Medium-to-Low Sparsity}
The typical utility of pruning is to produce a sparse yet competitive model that can benefit downstream applications in terms of efficiency without sacrificing much task accuracy. We hypothesize that PINS might also bring a regularization effect compared to vanilla fine-tuning under the medium-to-low sparsity regime. 

As shown in \tabref{table:mtl}, when specifying a medium-to-low sparsity, e.g., 50\%$\sim$30\%, our method can effectively play a role of regularization and improve model performance compared to vanilla fine-tuning. With half of the parameters being pruned, the sparse model produced by PINS outperforms fine-tuning by 
1 percentage point on the GLUE score. This observation suggests that appropriate pruning can effectively reduce variance without hurting model expressiveness.

\subsection{Ablation Study}
The self-regularization scheme is proposed and integrated into PINS to improve model generalization. Here we investigate the effectiveness of self-regularization by comparing it to the conventional knowledge distillation scheme and the classical empirical risk minimization scheme.

The pruning results of using the three different learning objectives on RTE, CoLA, and MRPC are listed in \tabref{table:ablation}. Pruning with PINS using classical empirical risk minimization still achieves performance better than existing baselines~(\tabref{table:glue}). Learning from a densely fine-tuned BERT$_{\text{base}}$ as the teacher does not always improve and sometime may even hurt performance. In contrast, our proposed self-regularization consistently boosts model performance, which echoes our theoretical justification in \secref{sec:srr}.

\begin{table}[t]
	\centering
	\small
	\begin{tabular}{c|ccc}
		\toprule
		$\mathcal{L}$                          & RTE  & CoLA & MRPC\\
		\midrule
		empirical risk                  & 70.9 & 55.4  &90.6 \\
%		\midrule
		w/ knowledge distillatiojn & 70.3  & 56.0 &90.6 \\
		w/ self-regularization     & \textbf{72.7} & \textbf{57.1} &90.9 \\
		\bottomrule
	\end{tabular}
\caption{Ablation Study with BERT$_{\text{base}}$ on the learning objective during iterative pruning at 80\% sparsity.}
\label{table:ablation}
\end{table}

\subsection{Analysis}
\begin{figure}[t]
	\centering
	\scalebox{0.236}{\includegraphics{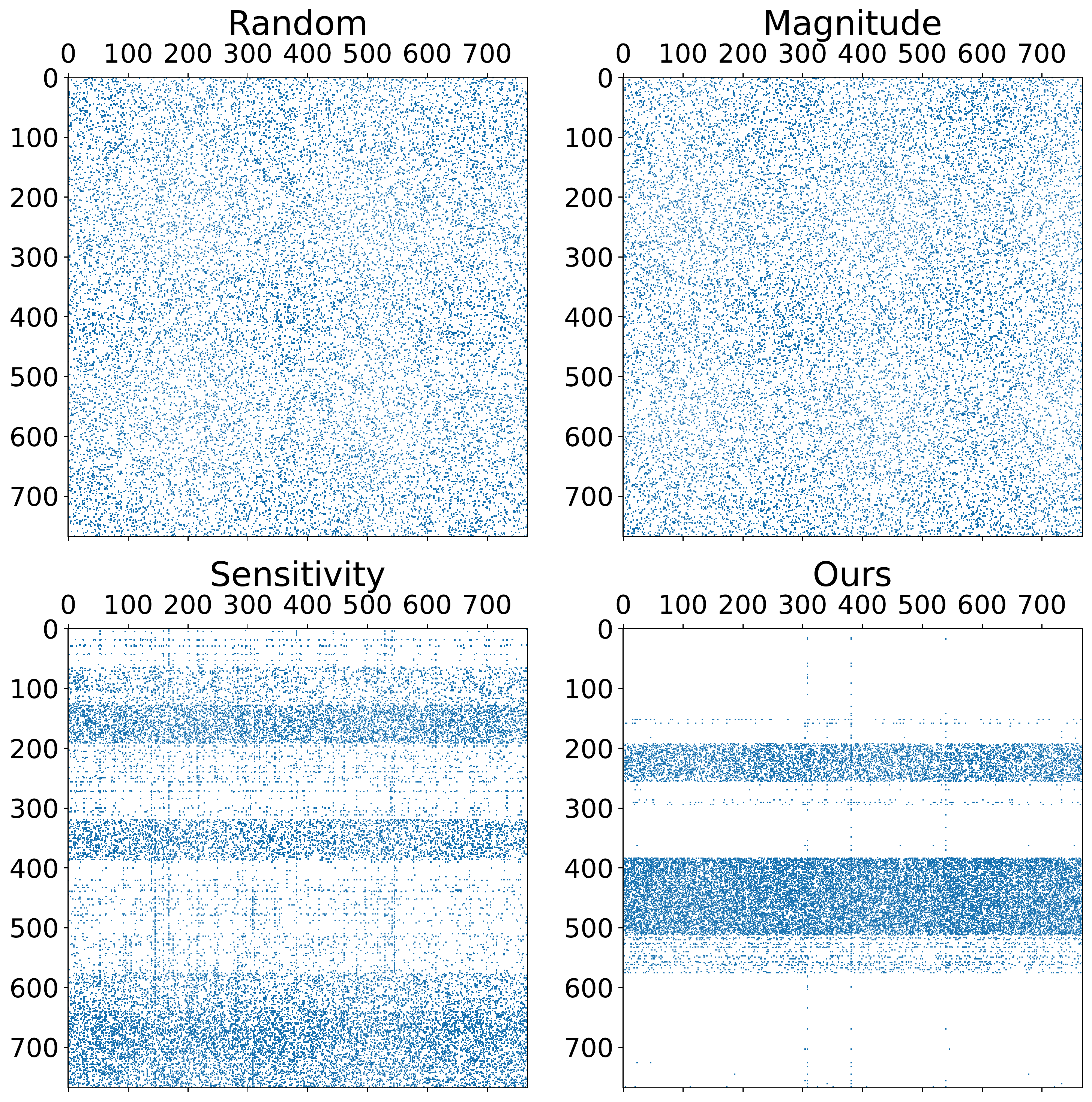}}
	\caption{Sparsity pattern~(80\%) of the same weight matrix in BERT$_{\text{base}}$ trained on SST-2. See Appendix \ref{sec:C} for more details on the matrix rank distribution.}
	\label{fig:sp}
\end{figure}

We provide an in-depth analysis of various importance criteria to uncover more valuable insights.
\paragraph{Sparsity pattern of weight matrices} We are interested in the sparsity pattern produced by different pruning criteria. To this end, we plot the remaining parameters' distribution of the same weight matrix in BERT$_{\text{base}}$ pruned via magnitude, sensitivity, and PINS in \figref{fig:sp}. We observe that magnitude-based pruning generates a sparsity pattern close to randomness. Sensitivity-based pruning produces a more structured pattern where the remaining parameters tend to occupy complete rows. Interestingly, the sparsity pattern produced by PINS exhibits the highest concentration on specific rows. This implies that the parameters contributing most to the end-task are preferably distributed in a structured way and PINS is more effective at extracting such patterns.

\begin{figure}[t]
	\centering
	\scalebox{0.31}{\includegraphics{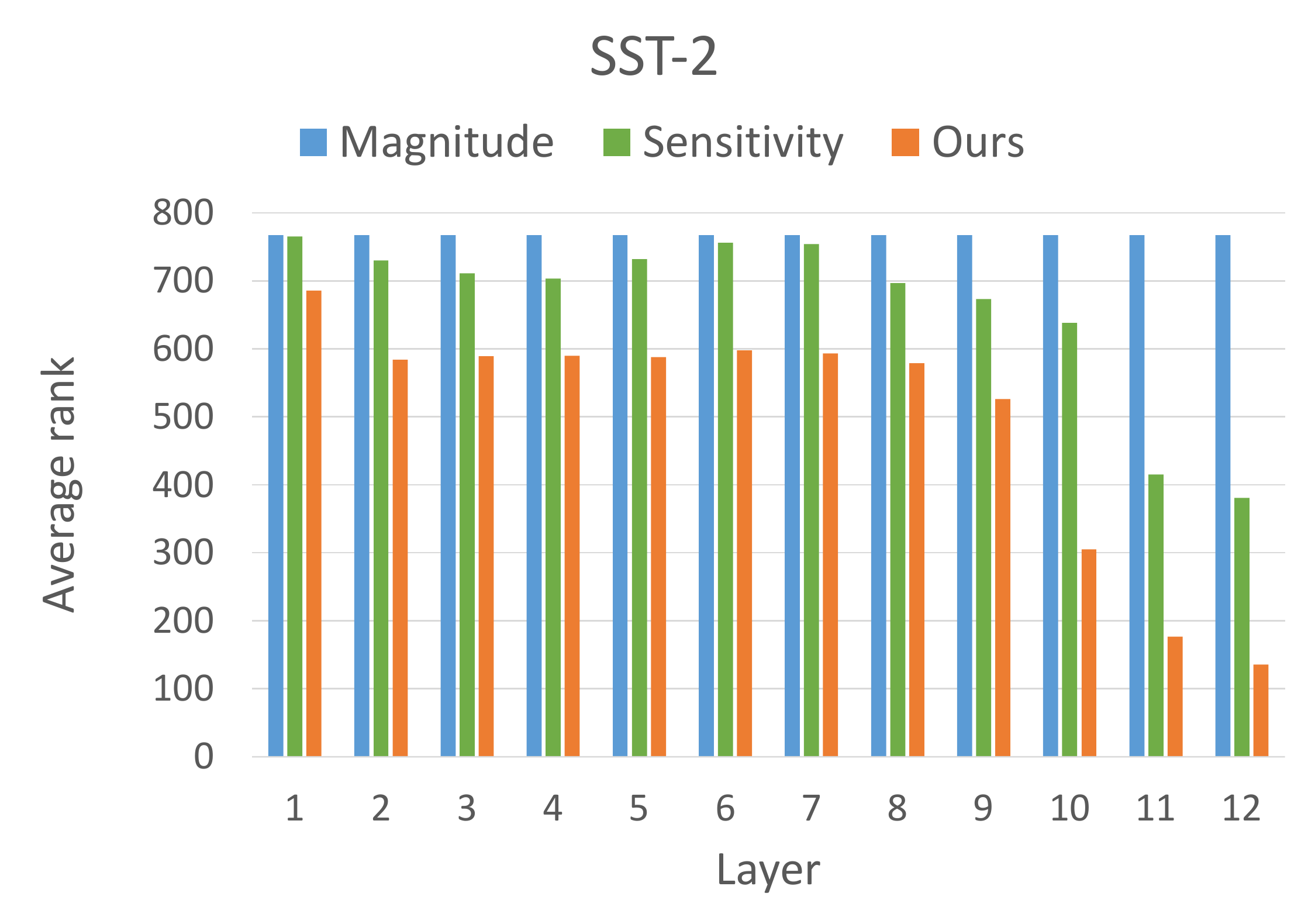}}
	\caption{Layerwise distribution of average matrix rank in BERT$_{\text{base}}$ pruned  at 80\% sparsity on  SST-2.}
	\label{fig:intersim}
\end{figure}

\paragraph{Layerwise rank distribution} 
The highly structured sparsity pattern generated by  PINS intrigues our interest to further analyze the intrinsic property of parameter matrices after pruning. Specifically, we inspect the matrix rank as it is usually associated with the complexity of matrix. To this end, we visualize the layerwise rank distribution of  BERT$_{\text{base}}$ pruned using different importance criteria on SST-2 dataset. As shown in \figref{fig:intersim}, magnitude pruning produces sparse matrices that are still near full-rank despite containing 80\% zeros. Sensitivity pruning tends to generate sparsity pattern with lower rank compared to magnitude pruning. Notably, model pruned by PINS shows consistently lower matrix rank than the other two criteria. This implies that PINS is more effective at identifying the low-dimensional task representation during adaptation, which is usually correlated with tighter generalization bounds~\cite{compressionbounds,intrinsicdimension}.

\begin{figure}[t]
	\centering
	\scalebox{0.31}{\includegraphics{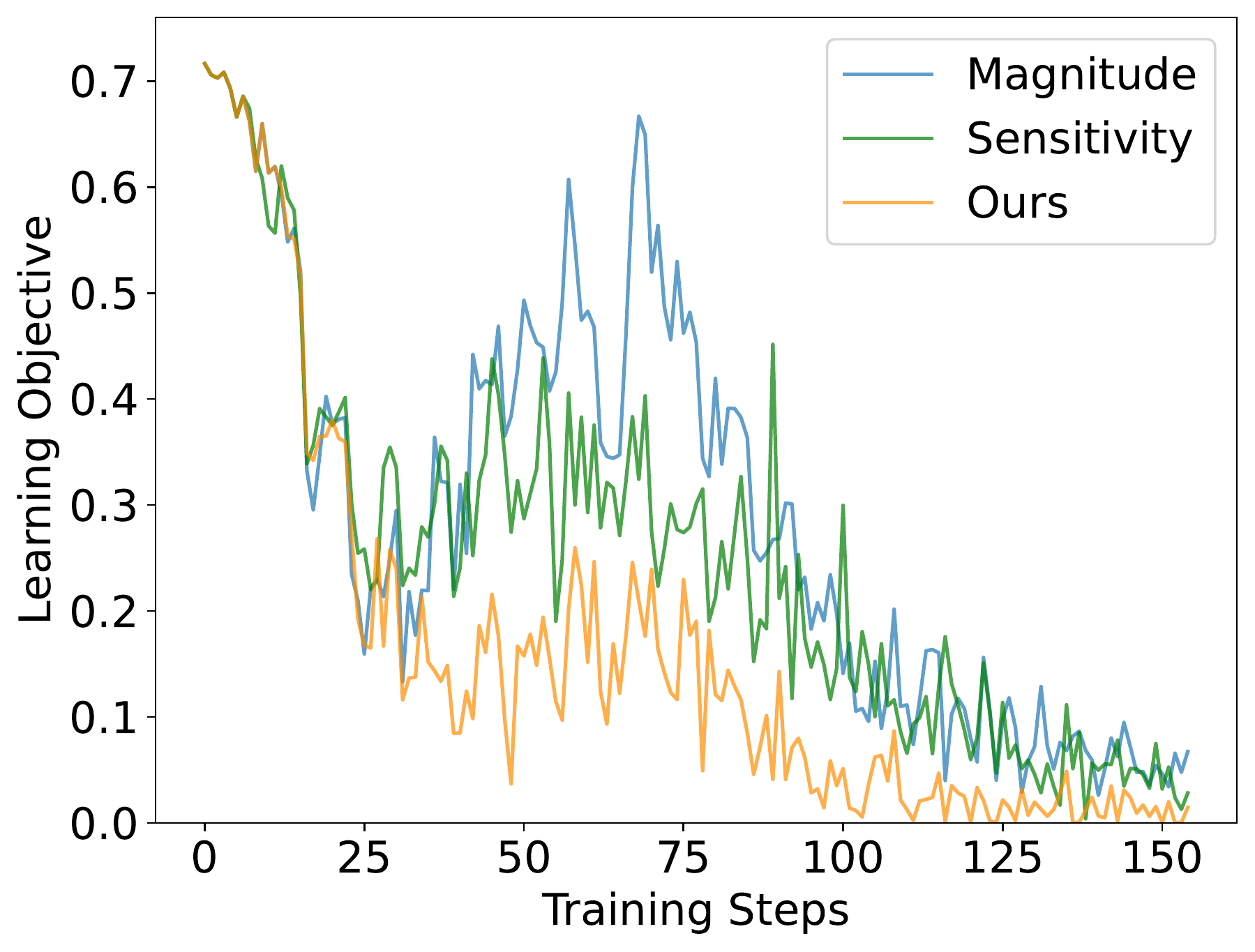}}
	\caption{Change of learning objective~(cross-entropy) during iterative pruning on SST-2.}
	\label{fig:lo}
\end{figure}

\paragraph{Empirical validation of importance criterion}
In \secref{sec:analysis} we prove that the pruning decision derived by our importance criterion is theoretically optimal. Here we empirically validate this point by visualizing the change of learning objective as pruning proceeds. \figref{fig:lo} illustrates that our importance criterion indeed leads to the most significant decrease in the learning objective compared to heuristical ones like magnitude and sensitivity.
%\KZ{If every step we take at PINS is optimal, then how come the orange line
%also bumps up sometimes, instead of monotonically downward?}
%\begin{figure}[t]
%	\centering
%	\scalebox{0.3}{\includegraphics{./figures/rank_dist_1.pdf}}
%	\caption{Layer-wise rank distribution of weight matrices after pruning. The maximum rank is 768 for BERT$_{\text{base}}$.}
%	\label{fig:rdist}
%\end{figure}

\begin{table}[t]
	\centering
	\small
	\begin{tabular}{c|cc|c}
		\toprule
		Sparsity & Time(s) & Storage(MB) & Acc. \\
		\midrule
		0\%      & 0.110~(1.0x)          & 340~(1.0x)  &  69.3\\
		80\%     & 0.041~(2.7x)          & 38~(8.9x) & 69.0 \\
		\bottomrule
	\end{tabular}
\caption{Practical time and storage efficiency gain on RTE with Deepsparse and CSR format. Inference is perform on Intel Xeon E5-2640 CPU with batch size 1.}
\label{table:eg}
\end{table}
\subsection{Efficiency Gain}
We can exploit the resulting high sparsity to attain practical efficiency gain on storage and inference speed. We first apply quantization  upon the pruned model and transform it into INT8 data type before saving it using Compressed Sparse Row~(CSR) format. We then leverage a sparsity-aware runtime~\cite{deepsparse} for accelerating inference. As shown in \tabref{table:eg}, on the RTE dataset, the disk space and inference time of BERT$_{\text{base}}$ pruned at 80\% sparsity can be reduced by roughly 8.9x and 2.7x respectively with negligible accuracy loss.

	\section{Conclusion}
We present PINS, a new iterative pruning method that hinges on a principled weight importance criterion to deliver the optimal stepwise pruning decision. Integrated with a self-regularization scheme tailored to pruning-during-adaptation, PINS allows for provably better generalization ability. Empirical experiments and analyses confirm the effectiveness of our method and shed further light on the different sparsity patterns produced by PINS and other existing methods.

\section*{Limitations}
Compared to the empirical risk minimization scheme, the introduced self-regularization scheme incurs certain overhead because each mini-batch of data will go through two models. For BERT$_{\text{base}}$ scale pre-trained language models, the additional memory overhead is about 27\% and the additional training time overhead is about 30\%. Nevertheless, once pruned, the sparsified model can enjoy considerable efficiency gains in terms of storage and inference time. Therefore, this is a trade-off that future practitioners might need to consider.

\section*{Acknowledgments}
This work was generously supported by the CMB Credit Card Center \& SJTU
joint research grant, and Meituan-SJTU joint research grant.

	\bibliography{paper}

\begin{thebibliography}{31}
\expandafter\ifx\csname natexlab\endcsname\relax\def\natexlab#1{#1}\fi

\bibitem[{Aghajanyan et~al.(2021)Aghajanyan, Gupta, and
  Zettlemoyer}]{intrinsicdimension}
Armen Aghajanyan, Sonal Gupta, and Luke Zettlemoyer. 2021.
\newblock \href {https://doi.org/10.18653/v1/2021.acl-long.568} {Intrinsic
  dimensionality explains the effectiveness of language model fine-tuning}.
\newblock In \emph{Proceedings of the 59th Annual Meeting of the Association
  for Computational Linguistics and the 11th International Joint Conference on
  Natural Language Processing (Volume 1: Long Papers)}, pages 7319--7328,
  Online. Association for Computational Linguistics.

\bibitem[{Arora et~al.(2018)Arora, Ge, Neyshabur, and
  Zhang}]{compressionbounds}
Sanjeev Arora, Rong Ge, Behnam Neyshabur, and Yi~Zhang. 2018.
\newblock Stronger generalization bounds for deep nets via a compression
  approach.
\newblock In \emph{International Conference on Machine Learning}, pages
  254--263. PMLR.

\bibitem[{Ben~Noach and Goldberg(2020)}]{svd}
Matan Ben~Noach and Yoav Goldberg. 2020.
\newblock \href {https://aclanthology.org/2020.aacl-main.88} {Compressing
  pre-trained language models by matrix decomposition}.
\newblock In \emph{Proceedings of the 1st Conference of the Asia-Pacific
  Chapter of the Association for Computational Linguistics and the 10th
  International Joint Conference on Natural Language Processing}, pages
  884--889, Suzhou, China. Association for Computational Linguistics.

\bibitem[{Clark et~al.(2020)Clark, Luong, Le, and Manning}]{electra}
Kevin Clark, Minh-Thang Luong, Quoc~V Le, and Christopher~D Manning. 2020.
\newblock \href {https://arxiv.org/abs/2003.10555} {Electra: Pre-training text
  encoders as discriminators rather than generators}.

\bibitem[{Devlin et~al.(2019)Devlin, Chang, Lee, and Toutanova}]{bert}
Jacob Devlin, Ming-Wei Chang, Kenton Lee, and Kristina Toutanova. 2019.
\newblock \href {https://doi.org/10.18653/v1/N19-1423} {{BERT}: Pre-training of
  deep bidirectional transformers for language understanding}.
\newblock In \emph{Proceedings of the 2019 Conference of the North {A}merican
  Chapter of the Association for Computational Linguistics: Human Language
  Technologies, Volume 1 (Long and Short Papers)}, pages 4171--4186,
  Minneapolis, Minnesota. Association for Computational Linguistics.

\bibitem[{Du{\v{s}}ek et~al.(2020)Du{\v{s}}ek, Novikova, and Rieser}]{e2e}
Ond\v{r}ej Du{\v{s}}ek, Jekaterina Novikova, and Verena Rieser. 2020.
\newblock \href {https://doi.org/10.1016/j.csl.2019.06.009} {Evaluating the
  {{State}}-of-the-{{Art}} of {{End}}-to-{{End Natural Language Generation}}:
  {{The E2E NLG Challenge}}}.
\newblock \emph{Computer Speech \& Language}, 59:123--156.

\bibitem[{Frankle and Carbin(2018)}]{oneshot2}
Jonathan Frankle and Michael Carbin. 2018.
\newblock \href {http://arxiv.org/abs/1803.03635} {The lottery ticket
  hypothesis: Training pruned neural networks}.
\newblock \emph{CoRR}, abs/1803.03635.

\bibitem[{Kurtz et~al.(2020)Kurtz, Kopinsky, Gelashvili, Matveev, Carr, Goin,
  Leiserson, Moore, Nell, Shavit, and Alistarh}]{deepsparse}
Mark Kurtz, Justin Kopinsky, Rati Gelashvili, Alexander Matveev, John Carr,
  Michael Goin, William Leiserson, Sage Moore, Bill Nell, Nir Shavit, and Dan
  Alistarh. 2020.
\newblock \href {http://proceedings.mlr.press/v119/kurtz20a.html} {Inducing and
  exploiting activation sparsity for fast inference on deep neural networks}.
\newblock In \emph{Proceedings of the 37th International Conference on Machine
  Learning}, volume 119 of \emph{Proceedings of Machine Learning Research},
  pages 5533--5543, Virtual. PMLR.

\bibitem[{Lan et~al.(2020)Lan, Chen, Goodman, Gimpel, Sharma, and
  Soricut}]{albert}
Zhenzhong Lan, Mingda Chen, Sebastian Goodman, Kevin Gimpel, Piyush Sharma, and
  Radu Soricut. 2020.
\newblock \href
  {http://dblp.uni-trier.de/db/conf/iclr/iclr2020.html#LanCGGSS20} {Albert: A
  lite bert for self-supervised learning of language representations.}
\newblock In \emph{ICLR}. OpenReview.net.

\bibitem[{Lee et~al.(2018)Lee, Ajanthan, and Torr}]{oneshot1}
Namhoon Lee, Thalaiyasingam Ajanthan, and Philip Torr. 2018.
\newblock Snip: Single-shot network pruning based on connection sensitivity.
\newblock In \emph{International Conference on Learning Representations}.

\bibitem[{Li et~al.(2022)Li, Luo, Tan, Wang, Huang, Li, and Bai}]{pst}
Yuchao Li, Fuli Luo, Chuanqi Tan, Mengdi Wang, Songfang Huang, Shen Li, and
  Junjie Bai. 2022.
\newblock \href {https://doi.org/10.24963/ijcai.2022/586} {Parameter-efficient
  sparsity for large language models fine-tuning}.
\newblock In \emph{Proceedings of the Thirty-First International Joint
  Conference on Artificial Intelligence, {IJCAI-22}}, pages 4223--4229.
  International Joint Conferences on Artificial Intelligence Organization.
\newblock Main Track.

\bibitem[{Liang et~al.(2021)Liang, Zuo, Chen, Jiang, Liu, He, Zhao, and
  Chen}]{superticket}
Chen Liang, Simiao Zuo, Minshuo Chen, Haoming Jiang, Xiaodong Liu, Pengcheng
  He, Tuo Zhao, and Weizhu Chen. 2021.
\newblock \href {https://doi.org/10.18653/v1/2021.acl-long.510} {Super tickets
  in pre-trained language models: From model compression to improving
  generalization}.
\newblock In \emph{Proceedings of the 59th Annual Meeting of the Association
  for Computational Linguistics and the 11th International Joint Conference on
  Natural Language Processing (Volume 1: Long Papers)}, pages 6524--6538,
  Online. Association for Computational Linguistics.

\bibitem[{Lopez-Paz et~al.(2015)Lopez-Paz, Bottou, Sch{\"o}lkopf, and
  Vapnik}]{difficulty1}
David Lopez-Paz, L{\'e}on Bottou, Bernhard Sch{\"o}lkopf, and Vladimir Vapnik.
  2015.
\newblock Unifying distillation and privileged information.
\newblock \emph{arXiv preprint arXiv:1511.03643}.

\bibitem[{Loshchilov and Hutter(2017)}]{adamw}
Ilya Loshchilov and Frank Hutter. 2017.
\newblock \href {http://arxiv.org/abs/1711.05101} {Fixing weight decay
  regularization in adam}.
\newblock \emph{CoRR}, abs/1711.05101.

\bibitem[{Louizos et~al.(2018)Louizos, Welling, and Kingma}]{l0}
Christos Louizos, Max Welling, and Diederik~P Kingma. 2018.
\newblock Learning sparse neural networks through $ l\_0 $ regularization.
\newblock \emph{arXiv preprint arXiv:1712.01312}.

\bibitem[{Mirzadeh et~al.(2019)Mirzadeh, Farajtabar, Li, and
  Ghasemzadeh}]{difficulty2}
Seyed{-}Iman Mirzadeh, Mehrdad Farajtabar, Ang Li, and Hassan Ghasemzadeh.
  2019.
\newblock \href {http://arxiv.org/abs/1902.03393} {Improved knowledge
  distillation via teacher assistant: Bridging the gap between student and
  teacher}.
\newblock \emph{CoRR}, abs/1902.03393.

\bibitem[{Radev et~al.(2020)Radev, Zhang, Rau, Sivaprasad, Hsieh, Rajani, Tang,
  Vyas, Verma, Krishna, Liu, Irwanto, Pan, Rahman, Zaidi, Mutuma, Tarabar,
  Gupta, Yu, Tan, Lin, Xiong, and Socher}]{dart}
Dragomir Radev, Rui Zhang, Amrit Rau, Abhinand Sivaprasad, Chiachun Hsieh,
  Nazneen~Fatema Rajani, Xiangru Tang, Aadit Vyas, Neha Verma, Pranav Krishna,
  Yangxiaokang Liu, Nadia Irwanto, Jessica Pan, Faiaz Rahman, Ahmad Zaidi,
  Murori Mutuma, Yasin Tarabar, Ankit Gupta, Tao Yu, Yi~Chern Tan, Xi~Victoria
  Lin, Caiming Xiong, and Richard Socher. 2020.
\newblock Dart: Open-domain structured data record to text generation.
\newblock \emph{arXiv preprint arXiv:2007.02871}.

\bibitem[{Radford et~al.(2018)Radford, Wu, Child, Luan, Amodei, and
  Sutskever}]{gpt2}
Alec Radford, Jeffrey Wu, Rewon Child, David Luan, Dario Amodei, and Ilya
  Sutskever. 2018.
\newblock \href
  {https://d4mucfpksywv.cloudfront.net/better-language-models/language-models.pdf}
  {Language models are unsupervised multitask learners}.

\bibitem[{Rajpurkar et~al.(2016)Rajpurkar, Zhang, Lopyrev, and Liang}]{squad}
Pranav Rajpurkar, Jian Zhang, Konstantin Lopyrev, and Percy Liang. 2016.
\newblock \href {http://arxiv.org/abs/1606.05250} {Squad: 100, 000+ questions
  for machine comprehension of text}.
\newblock \emph{CoRR}, abs/1606.05250.

\bibitem[{Renda et~al.(2020)Renda, Frankle, and Carbin}]{mag2}
Alex Renda, Jonathan Frankle, and Michael Carbin. 2020.
\newblock \href {http://arxiv.org/abs/2003.02389} {Comparing rewinding and
  fine-tuning in neural network pruning}.
\newblock \emph{CoRR}, abs/2003.02389.

\bibitem[{Sanh et~al.(2020)Sanh, Wolf, and Rush}]{movement}
Victor Sanh, Thomas Wolf, and Alexander Rush. 2020.
\newblock \href
  {https://proceedings.neurips.cc/paper/2020/file/eae15aabaa768ae4a5993a8a4f4fa6e4-Paper.pdf}
  {Movement pruning: Adaptive sparsity by fine-tuning}.
\newblock In \emph{Advances in Neural Information Processing Systems},
  volume~33, pages 20378--20389. Curran Associates, Inc.

\bibitem[{Shen et~al.(2019)Shen, Dong, Ye, Ma, Yao, Gholami, Mahoney, and
  Keutzer}]{eigen}
Sheng Shen, Zhen Dong, Jiayu Ye, Linjian Ma, Zhewei Yao, Amir Gholami,
  Michael~W. Mahoney, and Kurt Keutzer. 2019.
\newblock \href {http://arxiv.org/abs/1909.05840} {{Q-BERT:} hessian based
  ultra low precision quantization of {BERT}}.
\newblock \emph{CoRR}, abs/1909.05840.

\bibitem[{Sun et~al.(2019)Sun, Cheng, Gan, and Liu}]{pkd}
Siqi Sun, Yu~Cheng, Zhe Gan, and Jingjing Liu. 2019.
\newblock \href {http://arxiv.org/abs/1908.09355} {Patient knowledge
  distillation for {BERT} model compression}.
\newblock \emph{CoRR}, abs/1908.09355.

\bibitem[{Tjong Kim~Sang and De~Meulder(2003)}]{conll2003}
Erik~F. Tjong Kim~Sang and Fien De~Meulder. 2003.
\newblock \href {https://www.aclweb.org/anthology/W03-0419} {Introduction to
  the {C}o{NLL}-2003 shared task: Language-independent named entity
  recognition}.
\newblock In \emph{Proceedings of the Seventh Conference on Natural Language
  Learning at {HLT}-{NAACL} 2003}, pages 142--147.

\bibitem[{Vapnik(1998)}]{vc}
Vladimir Vapnik. 1998.
\newblock \emph{Statistical learning theory}.
\newblock Wiley.

\bibitem[{Vaswani et~al.(2017)Vaswani, Shazeer, Parmar, Uszkoreit, Jones,
  Gomez, Kaiser, and Polosukhin}]{transformer}
Ashish Vaswani, Noam Shazeer, Niki Parmar, Jakob Uszkoreit, Llion Jones,
  Aidan~N Gomez, \L~ukasz Kaiser, and Illia Polosukhin. 2017.
\newblock \href
  {https://proceedings.neurips.cc/paper/2017/file/3f5ee243547dee91fbd053c1c4a845aa-Paper.pdf}
  {Attention is all you need}.
\newblock In \emph{Advances in Neural Information Processing Systems},
  volume~30. Curran Associates, Inc.

\bibitem[{Wang et~al.(2018)Wang, Singh, Michael, Hill, Levy, and Bowman}]{glue}
Alex Wang, Amanpreet Singh, Julian Michael, Felix Hill, Omer Levy, and
  Samuel~R. Bowman. 2018.
\newblock \href {http://arxiv.org/abs/1804.07461} {{GLUE:} {A} multi-task
  benchmark and analysis platform for natural language understanding}.
\newblock \emph{CoRR}, abs/1804.07461.

\bibitem[{Wang et~al.(2020)Wang, Wei, Dong, Bao, Yang, and Zhou}]{minilm}
Wenhui Wang, Furu Wei, Li~Dong, Hangbo Bao, Nan Yang, and Ming Zhou. 2020.
\newblock \href {http://arxiv.org/abs/2002.10957} {Minilm: Deep self-attention
  distillation for task-agnostic compression of pre-trained transformers}.
\newblock \emph{CoRR}, abs/2002.10957.

\bibitem[{Zhang et~al.(2022)Zhang, Zuo, Liang, Bukharin, He, Chen, and
  Zhao}]{platon}
Qingru Zhang, Simiao Zuo, Chen Liang, Alexander Bukharin, Pengcheng He, Weizhu
  Chen, and Tuo Zhao. 2022.
\newblock Platon: Pruning large transformer models with upper confidence bound
  of weight importance.
\newblock In \emph{International Conference on Machine Learning}, pages
  26809--26823. PMLR.

\bibitem[{Zhou and Lampouras(2020)}]{webnlg}
Giulio Zhou and Gerasimos Lampouras. 2020.
\newblock \href {https://aclanthology.org/2020.webnlg-1.22} {{W}eb{NLG}
  challenge 2020: Language agnostic delexicalisation for multilingual
  {RDF}-to-text generation}.
\newblock In \emph{Proceedings of the 3rd International Workshop on Natural
  Language Generation from the Semantic Web (WebNLG+)}, pages 186--191, Dublin,
  Ireland (Virtual). Association for Computational Linguistics.

\bibitem[{Zhu and Gupta(2017)}]{gupta}
Michael Zhu and Suyog Gupta. 2017.
\newblock To prune, or not to prune: exploring the efficacy of pruning for
  model compression.
\newblock \emph{arXiv preprint arXiv:1710.01878}.

\end{thebibliography}
	\bibliographystyle{acl_natbib}
	\newpage
	\appendix

\section{Results with More PLMs on subset of GLUE}
\label{sec:A}
In addition the widely used BERT and GPT-2 models, we also perform pruning experiments upon other two pre-trained language models: Electra$_{\text{base}}$ and MiniLM$_{\text{12L-384H}}$ to further verify the effectiveness of our method. 

Due to computing resource constraint, we restrict our experiments on a subset of GLUE task, including RTE, CoLA and QNLI at 80\% and 90\% sparsity. We compare PINS against IMP and PLATON as two representative baselines for magnitude-based and sensitivity-based pruning methods. We fix the batch size as 32 and learning rate as 3e-5 similar to the BERT experiments. We illustrate the pruning results on \tabref{table:electra} and \tabref{table:minilm}. At both sparsity levels, PINS consistently outperforms IMP and PLATON on all three datasets, verifying the general effectiveness of PINS for language model pruning.
\begin{table}[h]
	\centering
	%		\normalsize
	\small
	\begin{tabular}{c|l|ccc}
		\toprule
		\multirow{2}{*}{Sparsity} & \multicolumn{1}{c|}{\multirow{2}{*}{Method}} & \multicolumn{1}{c}{\multirow{2}{*}{\begin{tabular}[c]{@{}c@{}}\textbf{RTE}\\ Acc\end{tabular}}} & \multicolumn{1}{c}{\multirow{2}{*}{\begin{tabular}[c]{@{}c@{}}\textbf{CoLA}\\ Mcc\end{tabular}}} & \multicolumn{1}{c}{\multirow{2}{*}{\begin{tabular}[c]{@{}c@{}}\textbf{QNLI}\\ Acc\end{tabular}}} \\
		& \multicolumn{1}{c|}{}                        & \multicolumn{1}{c}{}                                                                   & \multicolumn{1}{c}{}                                                                   & \multicolumn{1}{c}{}                                                                                                                              \\
		\midrule
		\multicolumn{1}{c|}{0\%}   & Fine-tune                                  & 73.0                                                                                  &58.5
		&91.5                                                                                                                                                                                                                                                                                 \\
		\midrule
		\multirow{3}{*}{80\%}     &IMP                                                &60.5                                                                                       &   21.6                                                                                     &   87.5                                                                                               \\
		&PLATON                                            &  68.2                                                                                      &  54.1                                                                                      &     89.8                                                                                           \\
		&PINS~(ours)                                             &   69.5                                                                                   &      54.4                                                                                 &  90.4                                                                                          \\
		\midrule
		\multirow{3}{*}{90\%}     &IMP                                              &   57.5                                                                                    &   14.1                                                                                     &   83.9                                                                                     \\
		& PLATON                                               &   63.1                                                                                   &  38.8                                                                                      &  88.0                                                                                        \\
		&PINS~(ours)                                             &   66.2                                                                                    &     44.8                                                                                  &   88.6                                                                                  \\
		\bottomrule                                     
	\end{tabular}
	\caption{Results with MiniLM$_{\text{12L-384H}}$ on the GLUE development set.}
	\label{table:electra}
\end{table}
\begin{table}[h]
	\centering
	%	\normalsize
	\small
	\begin{tabular}{c|l|ccc}
		\toprule
		\multirow{2}{*}{Sparsity} & \multicolumn{1}{c|}{\multirow{2}{*}{Method}} & \multicolumn{1}{c}{\multirow{2}{*}{\begin{tabular}[c]{@{}c@{}}\textbf{RTE}\\ Acc\end{tabular}}} & \multicolumn{1}{c}{\multirow{2}{*}{\begin{tabular}[c]{@{}c@{}}\textbf{CoLA}\\ Mcc\end{tabular}}} & \multicolumn{1}{c}{\multirow{2}{*}{\begin{tabular}[c]{@{}c@{}}\textbf{QNLI}\\ Acc\end{tabular}}} \\
		& \multicolumn{1}{c|}{}                        & \multicolumn{1}{c}{}                                                                   & \multicolumn{1}{c}{}                                                                   & \multicolumn{1}{c}{}                                                                                                                                   \\
		\midrule
		\multicolumn{1}{c|}{0\%}   & Fine-tune                                  &  81.9                                                                                 &69.0
		&  93.1                                                                                                                                                                   \\
		\midrule
		\multirow{3}{*}{80\%}     &IMP                                                & 59.9                                                                                      &    11.2                                                                                    &  87.5                                                                                          \\
		&PLATON                                            &  73.6                                                                                      &    60.0                                                                                    & 91.0                                                                                        \\
		&PINS~(ours)                                             &    75.5                                                                                  &      63.7                                                                                 &  92.0                                                                                     \\
		\midrule
		\multirow{3}{*}{90\%}     &IMP                                              &   52.9                                                                                    &0.0                                                                                        &    83.0                                                                                 \\
		& PLATON                                               &  69.9                                                                                    &    48.0                                                                                    &   89.7                                                                                   \\
		&PINS~(ours)                                             &  72.3                                                                                     &     49.2                                                                                  &      90.2                                                                                \\
		\bottomrule                                     
	\end{tabular}
	\caption{Results with Electra$_{\text{base}}$ on the GLUE development set.}
	\label{table:minilm}
\end{table}

\section{Proof of Theorem 1}
\label{sec:B}
%Proof of Theorem 1:
\begin{proof}
	Let $t_i$ and $t_{j}$ where $t_{i}\geq t_{j}$ denote the time steps at which two different checkpoints are saved; Let $R(f_{\bm{\theta}^{(t\leftarrow t_i)}})$ and $R(f_{\bm{\theta}^{(t\leftarrow t_j)}})$ denote the expected generalization error of models learned from  $f_{\bm{\theta}^{(t_i)}}$ and $f_{\bm{\theta}^{(t_j)}}$; Let $n$ denotes the size of training data; $|\cdot|_{\text{C}}$ denotes a capacity measure like VC-dimension for function class $\mathcal{F}_{\bm{\theta}}$. Based on previous expositions on VC theory, the following asymptotic generalization bound holds:
	\begin{align}\nonumber
        & R(f_{\bm{\theta}^{(t\leftarrow t_i)}})=R(f_{\bm{\theta}^{(t\leftarrow t_i)}})-R(f_{\bm{\theta}^{(t_i)}}) \\ \nonumber
        & +R(f_{\bm{\theta}^{(t_i)}}) \\ \nonumber
		&\leq O(\frac{|\mathcal{F}_{\bm{\theta}^{(t)}}|_{\text{C}}}{n^{\alpha_{i}}})+ \epsilon_{t,t_i} + R(f_{\bm{\theta}^{(t_i)}}) \\ \nonumber 
		&=  \underbrace{O(\frac{|\mathcal{F}_{\bm{\theta}^{(t)}}|_{\text{C}}}{n^{\alpha_{i}}}) + \underset{f_{\bm{\theta}^{(t)}}\in \mathcal{F}_{\bm{\theta}^{(t\leftarrow t_i)}}}{\inf}R(f_{\bm{\theta}^{(t)}})}_{bound(f_{\bm{\theta}^{(t\leftarrow t_i)}})} \\ \nonumber
        &R(f_{\bm{\theta}^{(t\leftarrow t_j)}})=R(f_{\bm{\theta}^{(t\leftarrow t_j)}})-R(f_{\bm{\theta}^{(t_j)}})\\ \nonumber 
        &+R(f_{\bm{\theta}^{(t_j)}}) \\ \nonumber
		&\leq O(\frac{|\mathcal{F}_{\bm{\theta}^{(t)}}|_{\text{C}}}{n^{\alpha_{j}}})+ \epsilon_{t,t_j} + R(f_{\bm{\theta}^{(t_j)}}) \\ \nonumber 
		&=  \underbrace{O(\frac{|\mathcal{F}_{\bm{\theta}^{(t)}}|_{\text{C}}}{n^{\alpha_{j}}}) + \underset{f_{\bm{\theta}^{(t)}}\in \mathcal{F}_{\bm{\theta}^{(t\leftarrow t_j)}}}{\inf}R(f_{\bm{\theta}^{(t)}})}_{bound(f_{\bm{\theta}^{(t\leftarrow t_j)}})}
	\end{align}
	where $\epsilon_{t,ti}$ is the approximation error of function class $\mathcal{F}_{\bm{\theta}^{(t\leftarrow t_i)}}$ with respect to $f_{\bm{\theta}^{(t_i)}}$. $\epsilon_{t,tj}$ is defined in analogy.
	Because: (1) $\bm{\theta}^{(t_i)}$ is a later checkpoint with higher sparsity than $\bm{\theta}^{(t_j)}$, we have the learning speed $1\geq \alpha_{i}\geq \alpha_{j}\geq \frac{1}{2}$; (2) $f_{\bm{\theta}^{(t_i)}}$ has lower generalization error than $f_{\bm{\theta}^{(t_j)}}$, we have the following inequality holds with high probability:
	\begin{align}\nonumber
		bound(f_{\bm{\theta}^{(t\leftarrow t_i)}}) \leq bound(f_{\bm{\theta}^{(t\leftarrow t_j)}})
	\end{align}
\end{proof}

\begin{figure*}[t]
	\centering
	\scalebox{0.48}{\includegraphics{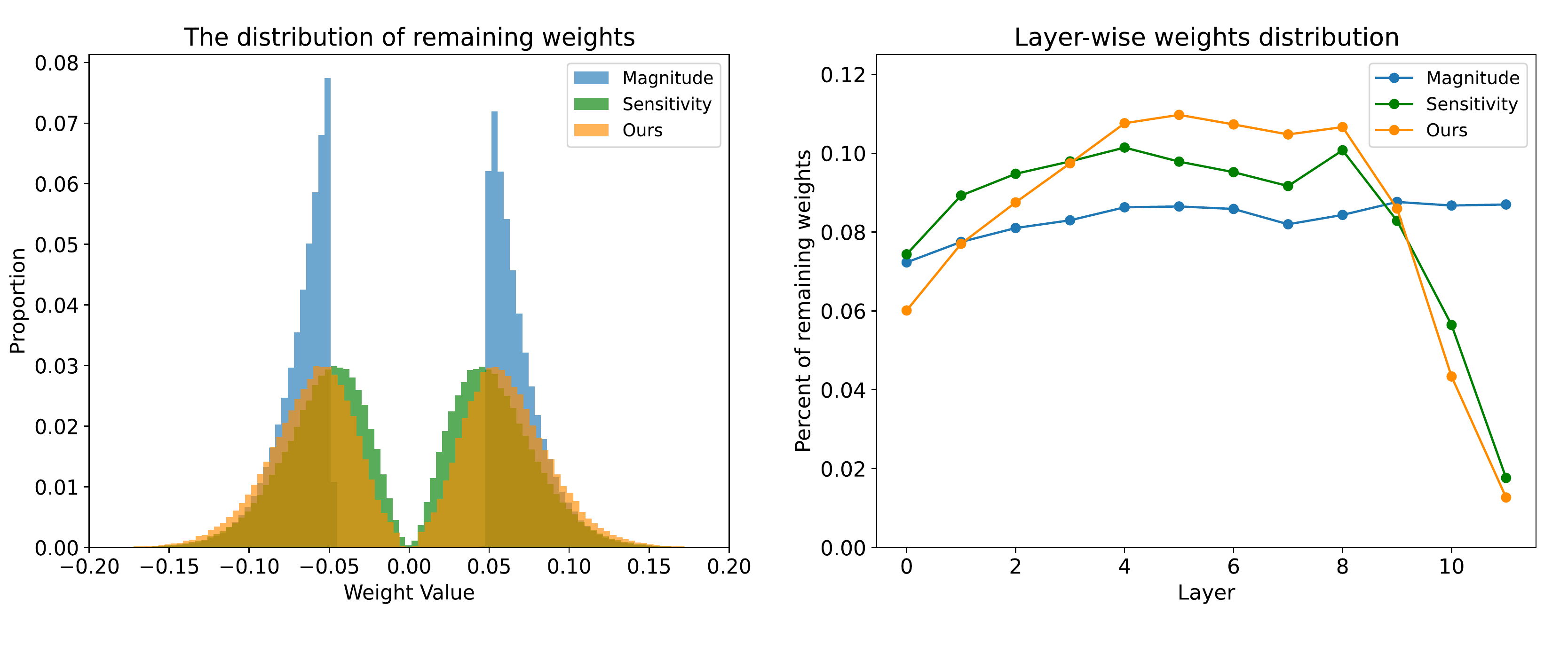}}
	\caption{Weight distributions of BERT$_{\text{base}}$ pruned using different importance criteria on RTE dataset. Left figure shows the value distribution and the right figure shows how remaining parameters are distributed at different model layers.}
	\label{fig:intersim}
\end{figure*}
\begin{figure}[t]
	\centering
	\scalebox{0.355}{\includegraphics{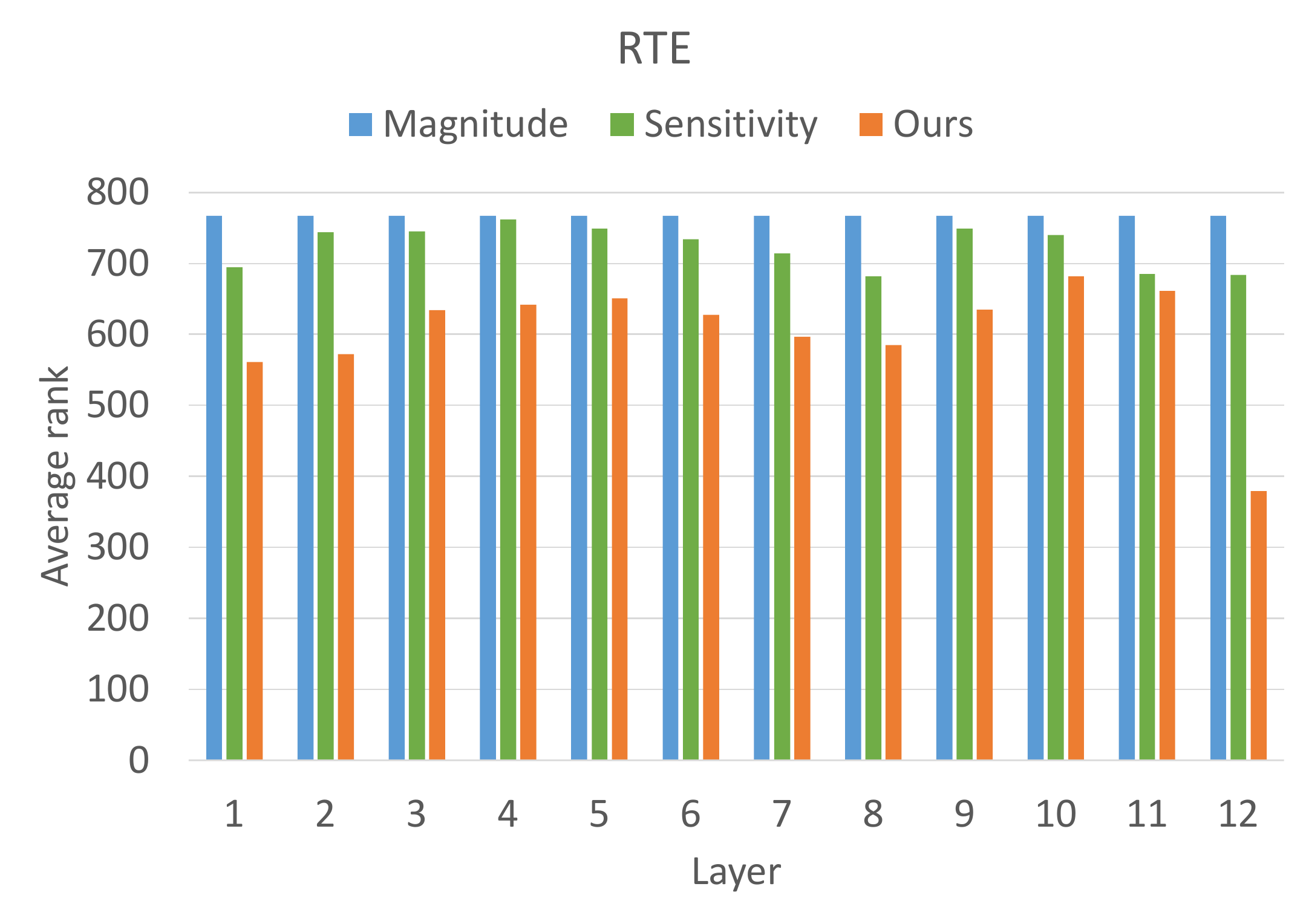}}
	\caption{Layerwise rank distribution of BERT$_{\text{base}}$ pruning using different importance criteria on RTE dataset.}
	\label{fig:rank_dist}
\end{figure}

\section{More Post-pruning Analyses}
\label{sec:C}
This section presents more visualized analyses of models sparsified by different pruning methods. 

\figref{fig:rank_dist} shows the layerwise rank distribution of BERT$_{\text{base}}$ pruned using different importance criteria on the RTE dataset. The observation here is similar to what is discussed in the main body of the paper: PINS exhibits the lowest average matrix rank in the sparsified model compared to the other two criteria.

\figref{fig:intersim} illustrates the weight distribution of BERT$_{\text{base}}$ pruning using different importance criteria. From the left figure we can see that magnitude-based pruning tends to keep parameters with high absolute values, which is expected based on its definition. Sensitivity and PINS produce similar weight value distribution mainly because the two methods both contain the $g\theta$ term in their importance calculation. Despite the similarity, we can still observe that PINS produces smoother distribution than sensitivity and covers more weights with larger absolute values.

The right figure shows the layerwise distribution of remaining parameters after pruning. A clear trend is that PINS tends to retain more parameters in the middle layers~(4-7), which also coincided with the inter-model sparsity pattern analysis in the main body of our paper. Both sensitivity and PINS remove a large proportion of parameters in the top layers~(10-12) while magnitude-based pruning has no preference for model layers.

\section{Sparsity Scheduler}
The proportion of remaining weights is controlled by the sparsity scheduler, here  we adopt the commonly used  cubic sparsity schedule to progressively reach target sparsity, i.e., $r^{(t)}$ at time step $t$ within the maximum time steps $T$ is given by:
\begin{align}
	%	v^{(t)}=
	\begin{cases} 
		r_i & t\in [0, t_i) \\
		r_f+(r_i-r_f)(\frac{T-t_{f}-t}{T-t_f-t_i})^3 & t\in[t_i, T-t_f) \\
		r_f  & \text{otherwise}  
	\end{cases}
\end{align}
\label{eq:prune}
where $r_i=1.0$, $r_f$ is the final percent of remained parameters, $t_i$ and $t_f$ are the warmup and cool-down steps.

\section{Accelerating Inference and Reducing Storage}
We attain practical efficiency gain in terms of inference time and disk storage space using different sets of off-the-shelf techniques. Specifically, we use DeepSparse\footnote{\url{https://github.com/neuralmagic/deepsparse}}, a sparsity-aware inference runtime to accelerate inference of sparse model on CPUs. We also utilize the Pytorch built-in quantization function\footnote{\url{https://pytorch.org/docs/stable/quantization.html}} and Compressed Sparse Row~(CSR) format\footnote{\url{https://github.com/huggingface/block_movement_pruning/blob/master/Saving_PruneBERT.ipynb}} to achieve a much smaller disk space requirement.

%	\clearpage
%	\input{appendix}
\end{document}